\documentclass[peerreview, draftcls, onecolumn]{IEEEtran}
\usepackage{cite, graphicx, amsmath, amsfonts, amssymb, array, latexsym,subfigure, url,multirow}

\newtheorem{theorem}{Proposition}[section]

\newcommand{\MS}{{{\rm SNL}_{s,t,k}^{(1)}}}
\newcommand{\MR}{{{\rm SNL}_{s,t,k}^{(2)}}}
\newcommand{\CSMS}{{{\rm SNL}_{s,t,k}^{(3)}}}
\newcommand{\CSMR}{{{\rm SNL}_{s,t,k}^{(4)}}}

\title{A New Similarity Measure for Non-Local Means Filtering of MRI Images}
\author{Sudipto Dolui, Alan Kuurstra, Iv\'{a}n C. Salgado Patarroyo and Oleg V. Michailovich %
\thanks{All authors are with the Department of Electrical and Computer Engineering at the University of Waterloo (Ontario, Canada). This research was supported by a Discovery grant from NSERC -- The Natural Sciences and Engineering Research Council of Canada. Information on various NSERC activities and programs can be obtained from {\tt http://www.nserc.ca}.}}

\begin{document}
\maketitle

\begin{abstract}
Magnetic resonance imaging (MRI) is a principal modality of modern medical imaging, which provides a wide spectrum of useful diagnostic contrasts, both anatomical and functional in nature. Like many alternative imaging modalities, however, some specific realizations of MRI offer a trade-off in terms of acquisition time, spatial/temporal resolution and signal-to-noise ratio (SNR). Thus, for instance, increasing the time efficiency of MRI often comes at the expense of reduced SNR. This, in turn, necessitates the use of post-processing tools for noise rejection, which makes image de-noising an indispensable component of computer assistance diagnosis. In the field of MRI, a multitude of image de-noising methods have been proposed hitherto. In this paper, the application of a particular class of de-noising algorithms -- known as non-local mean (NLM) filters -- is investigated. Such filters have been recently applied for MRI data enhancement and they have been shown to provide more accurate results as compared to many alternative de-noising algorithms. Unfortunately, virtually all existing methods for NLM filtering have been derived under the assumption of additive white Gaussian (AWG) noise contamination. Since this assumption is known to fail at low values of SNR, an alternative formulation of NLM filtering is required, which would take into consideration the correct Rician statistics of MRI noise. Accordingly, the contribution of the present paper is two-fold. First, it points out some principal disadvantages of the earlier methods of NLM filtering of MRI images and suggests means to rectify them. Second, the paper introduces a new similarity measure for NLM filtering of MRI Images, which is derived under {\it bona fide} statistical assumptions and results in more accurate reconstruction of MR scans as compared to alternative NLM approaches. Finally, the utility and viability of the proposed method is demonstrated through a series of numerical experiments using both {\it in silico} and {\it in vivo} MRI data.
\end{abstract}


\IEEEpeerreviewmaketitle

\section{Introduction} \label{Introduction}
Magnetic resonance imaging (MRI) is considered to be one of the most advanced modalities of modern medical imaging, which excels in providing a wide spectrum of useful diagnostic contrasts \cite{Wright97}. Since the latter constitute an intensity-coded representation of biological properties of studied tissues/organs, the precision with which a contrast represents its associated biological property plays a decisive role in tissue characterization and early diagnosis. This fact establishes the value of post-processing techniques which aim at improving the signal-to-noise ratio (SNR) of diagnostic MR images, while preserving the integrity and consistency of their anatomical content.     

Unfortunately, in virtually all realizations of MRI, attaining higher spatial resolution entails using longer acquisition times. Apart from being highly undesirable from the perspective of patients' comfort and compliance, longer acquisition times lead to motion-related artifacts, which are the main foe of cardiac and diffusion MRI \cite{Mcveigh85, Gudbjartsson95, Macovski96, Michailovich:2010xq}. On the other hand, the reduction of acquisition time results in a loss of the spatial resolution as well as in an amplification of measurement noises. The latter tend to obscure and mask diagnostically relevant details of MR scans, thereby necessitating the application of efficient and reliable tools of image de-noising \cite{Henkelman85}.   

The current arsenal of image de-noising methods used in MRI is immense, which makes their fair classification a non-trivial task. For this reason, only three groups of de-noising methods which are germane to the present developments are mentioned below, while the reader is referred to the references therein for a more comprehensive literature review. In particular, the first group of de-noising algorithms for MRI encompasses variational methods, which are implemented through the solution of partial differential equations (PDE) \cite{Gerig92, Lysaker03, Fan03, Basu06}. Thus, for example, \cite{Gerig92} suggests an adaptation of the classical anisotropic diffusion filter of \cite{PeronaMalik90} for noise reduction and enhancement of object boundaries in MRI. On the other hand, the de-noising method of \cite{Lysaker03} is based on minimization of an original cost functional, whose associated gradient flow has the form of a fourth-order PDE. In \cite{Fan03}, information from both the body coil image and surface coil image are incorporated in the form of data fidelity constraints. Finally, \cite{Basu06} introduces a {\it maximum-a-posteriori} (MAP) technique using a Rician noise model in combination with spatial regularization.

A different group of de-noising methods takes advantage of the sparsifying properties of certain linear transforms \cite {Weaver91, Healy92, Xu94, Hilton96, Nowak99, Wood99, Zaroubi00, Pizurica03, Bao03, Anand10} (for a comprehensive review of such methods, the reader is also referred to \cite{Pizurica06}). Thus, for instance, the method of \cite{Nowak99} is based on wavelet thresholding applied to squared-amplitude MR images, supplemented by ``unbiasing" of the scaling coefficients to account for the non-central chi-square distribution statistics. A different (robust) shrinkage scheme in the domain of a wavelet transform is proposed in \cite{Pizurica03}. Using a different line of arguments, the wavelet de-noising method of \cite{Zaroubi00} is applied to complex-valued MR images. Finally, in \cite{Anand10}, the MR images are enhanced by means of a wavelet-domain bilateral filter.

A third group of image de-noising algorithms is based on the concept of non-local means (NLM) filtering, which was originally proposed in \cite{BuadesImage05, BuadesReview05, Buades08}, with its later improvements reported in \cite{Dabov07,Boulanger07}. As a general rule, NLM filters estimate a noise-free intensity of a given pixel (source pixel) as a weighted (linear) combination of the rest of the image pixels (target pixels). Here, the weights of the linear combination are determined based on a {\em similarity measure} (SM) between the neighbourhoods of the target and  source pixels. As a result, the performance of an NLM filter is largely determined by the optimality of a chosen SM with respect to the properties of the image to be enhanced as well as those of the measurement noise. Thus, for example, under the conditions of additive white Gaussian (AWG) noise contamination, the above-referred NLM filters have been shown to outperform many variational and wavelet-based filters in terms of noise removal and the quality of edge preservation.    

Motivated by the success of NLM filtering in general image processing, the works in \cite{Daessle07, Coupe08} have extended the Gaussian-mode NLM filters to MR imagery. Additional reports on the subject also include \cite{Manjon08, Daessle08, ManjonAdaptive10}, where the filters are applied to MR images, followed by subtracting an estimation bias from the results thus obtained. 

Central to the main subject of the present work is the fact that the Rician statistics of MR images converges to Gaussian as the SNR of the images goes to infinity \cite{Gudbjartsson95}. For sufficiently high values of SNR, therefore, applying the Gaussian-mode NLM filters seem to be well justified \cite{Daessle08}. However, for relatively low values of SNR (as in the case with, e.g., diffusion weighted imaging), the Gaussian model ceases to be legitimate, and as a result, the NLM weights optimal for the Gaussian setting become sub-optimal. This fact suggests a need for an SM which remains optimal for a wide range of the SNR values. Accordingly, deriving such an SM constitutes the main objective of the present paper. To this end, we start with an analysis of the NLM filters recently proposed in \cite{Deledalle09, Teuber11} and underline some of their properties which should be avoided in the case of MR image de-noising. Subsequently, based on the results of \cite{Deledalle09, Teuber11}, we propose a new formulation of the SM and its associated weights, and demonstrate its usefulness and viability through a series of experiments using both {\it in silico} and {\it in vivo} MRI data.

Table \ref{T1} summarizes the main abbreviations and notations used in the paper, whose remainder is organized as follows. Section \ref{Sec:ImageModel} provides some necessary details on the image formation model of MR images and their noise statistics. Sections \ref{Sec:NLMeans} and \ref{Sec:PossibleApproach} describe a number of principal approaches to NLM filtering and point out some of their problematic aspects in relation to an MRI setting. A new SM and the closed-form expressions for its associated weights are derived in Section \ref{Sec:Proposed}, while Section \ref{Sec:BiasRemoval} details a method for applying these proposed weights to the noisy pixels of data images. Section \ref{Sec:Results} compares the performance of the proposed algorithm with that of some alternative methods using both {\it in silico} and {\it in vivo} MRI data. Finally, the main results and conclusions of the paper are recapitulated in Section \ref{Sec:Discussion}.

\begin{table}\label{T1}
\centering
\caption{List of Notations and Abbreviations}
\label{Table:1}
\begin{tabular*}{0.85\textwidth}{ c | l | c}
\hline
Notations and Abbreviations & Meaning & Formula (if applicable)\\
\hline\hline
NLM  	& Non-local means 					& - \\\hline
NCCS 	& Non-central chi square 				& - \\\hline
SM 		& Similarity measure 				& - \\\hline
SNL 	& Similarity measure for NLM 				& - \\\hline
SSM 	& Subtractive similarity measure		& \eqref{eq:AddMeasure} \\\hline 
RSM 	& Rational similarity measure 			& \eqref{OM5} \\\hline 
CSM 	& Correlation similarity measure		& - \\\hline
$\MS$ 	& Subtractive SM for NCCS distribution 	& \eqref{eq:SubtractiveNCCS}\\\hline
$\MR$ 	& Rational SM for Rician distribution 	& \eqref{eq:RatioRice}\\\hline 
$\CSMS$ & Subtractive CSM for NCCS distribution 	& \eqref{eq:CorrNCCS}\\\hline
$\CSMR$ & Rational CSM for Rician distribution 	& \eqref{eq:CorrRice}\\\hline 
\end{tabular*}
\end{table}

\section{Image Formation Model and Noise Statistics}\label{Sec:ImageModel}
\begin{figure}[ht]
\centering
\subfigure[]{
   \includegraphics[width=0.45\textwidth]{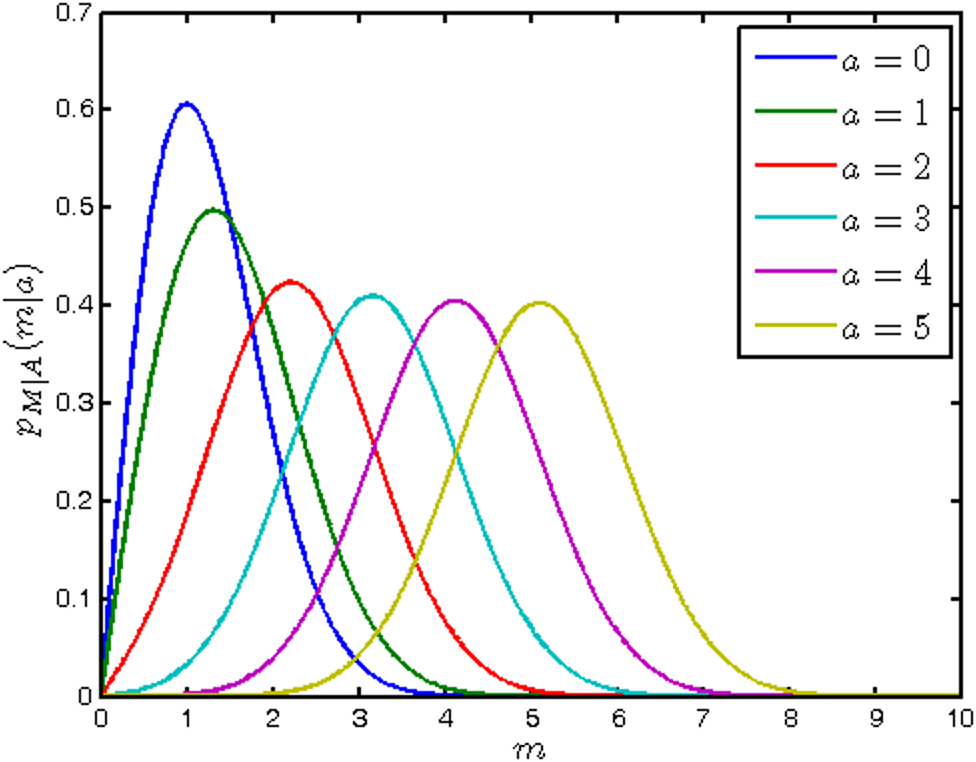}
   \label{Fig:Rice}
 }
 \subfigure[]{
   \includegraphics[width=0.45\textwidth]{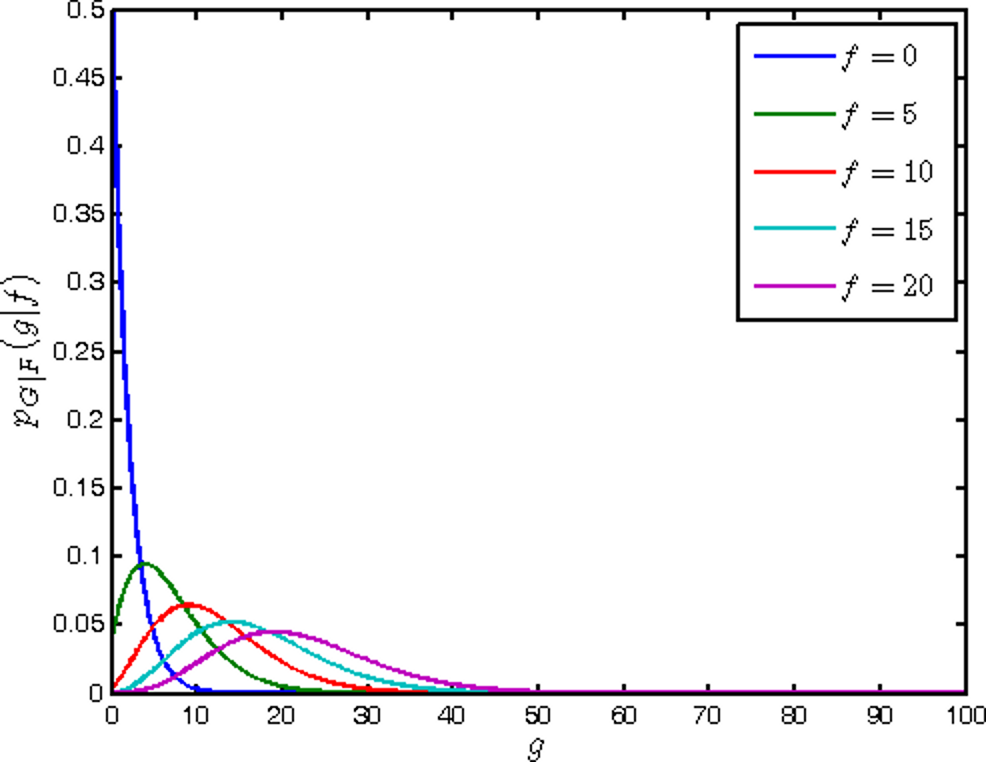}
   \label{Fig:NCCS}
 }
\label{fig:RiceNCCS}
\caption{(a) Rician pdf's corresponding to different values of $A$ and $\sigma = 1$ in \eqref{OM1}; (b) Non-central chi square distribution corresponding to different values of $F$ in \eqref{OM11}.}
\end{figure}

MR images are standardly acquired in the Fourier domain (i.e., the $k$-space), followed by the procedures of frequency demodulation and inverse transformation, which result in corresponding complex-valued images, whose magnitude is subsequently displayed \cite{Macovski96}. In this case, if the frequency-domain data is contaminated by zero-mean AWG noise, the complex amplitude $M$ of the noisy observation $A \exp\{ \imath \alpha\} + N$, with $N = N_r + \imath N_i$ , is given by 
\begin{equation}\label{eq:M}
M=\sqrt{(A\cos\alpha+N_r)^2+(A\sin\alpha+N_i)^2},
\end{equation}
where $A$ stands for the true image amplitude, while $N_r$ and $N_i$ are mutually independent AWG noises of standard deviation $\sigma$, and $\alpha \in [0, 2\pi )$ is an arbitrary phase shift. In this case, $M$ can be shown to follow the Rician conditional distribution model that is given by\footnote{Here and hereafter, we use the standard statistical formalism for denoting random variables and their associated realizations by capital letters and their lower-case counterparts, respectively.} \cite{Gudbjartsson95, Macovski96}
\begin{equation}\label{OM1}
p_{M|A}(m|a)=\begin{cases}\frac{m}{\sigma^2}\exp\left\{-\frac{a^2+m^2}{2\sigma^2}\right\}I_0\left(\frac{am}{\sigma^2}\right),&m\geq 0\\
0, & \mbox{otherwise}.
\end{cases} 
\end{equation}
where $I_0$ denotes the $0^{th}$-order modified Bessel function of the first kind. Figure \ref{Fig:Rice} depicts several typical shapes of $p_{M|A}(m|a)$ corresponding to a range of the values of $A$ and $\sigma=1$. As can be seen from the figure, for $A > 3 \sigma$, the Rician probability density function starts closely resembling that of a Gaussian random variable \cite{Gudbjartsson95}). However, for lower values of $A$, the density $p_{M|A}(m|a)$ becomes more asymmetric and protrudently heavy-tailed. Specifically, for $A=0$, $M$ follows a Rayleigh distribution model.

The Rician nature of $p_{M|A}$ in \eqref{OM1} renders impractical a straightforward application of many filtering strategies. This is because of the highly-nonlinear relation between the expectation $\mathcal{E}\{M\}$ of $M$ and $A$. Specifically, 
\begin{equation}
\mathcal{E}\{M\}=\sigma\sqrt{\pi/2}L_{1/2}(-A^2/2\sigma^2),
\end{equation}
where $L_v(x)$ denotes a Laguerre polynomial which, for $v=1/2$, is given by
\begin{equation}
L_{1/2}(x)=e^{x/2}\left[(1-x)I_0\left(-\frac{x}{2}\right)-xI_1\left(-\frac{x}{2}\right)\right].
\end{equation}

At the same time, a normalized version $G = (M/\sigma)^2$ of the squared magnitude $M^2$ can be shown to be distributed according to a {\em  non-central chi square} (NCCS) distribution with two degrees of freedom and parameter $F = (A / \sigma)^2$, whose conditional density is given by
\begin{equation}\label{OM11}
p_{G|F}(g|f)=\begin{cases}\frac{1}{2}e^{-(g+f)/2}I_0(\sqrt{fg}),&g\geq 0\\
0, & \mbox{otherwise},
\end{cases}
\end{equation}
where $f \in \mathbb{R}^+$. Figure \ref{Fig:NCCS} shows a number of typical shapes of $p_{G | F}$ corresponding to a set of different values of $F$. A better understanding of this figure can be derived from the fact that \eqref{eq:M} suggests
\begin{equation}\label{eq:G}
G=F+2\sqrt{F}\xi+\eta,
\end{equation}
where $\xi : =(N_r\cos\alpha+N_i\sin\alpha)/\sigma$ and $\eta : =(N_r^2+N_i^2)/\sigma^2$. Thus $G$ can be viewed as a noisy version of $F$, where the noise has both additive and multiplicative components. Specifically, it should be noted that $\xi$ obeys a normal distribution with zero mean and unit variance, while $\eta$ follows an exponential distribution with its mean and variance equal to $2$ and $4$, respectively. Moreover, the expectation of $G$ now has a very simple relation to $F$, which is given by 
\begin{equation}\label{eq:Eg}
\mathcal{E}\{G\}=F + 2.
\end{equation}
It is the simplicity of \eqref{eq:Eg} which has been a principal impetus for the development of various de-noising methods, which have been applied to the squared magnitude $G$, rather than to its original value $M$. In one way or another, all these methods aim at recovering a close approximation of the average value $\mathcal{E}\{G\}$, followed by the estimation of $F$ through the subtraction of the global bias of 2. Note that, once an estimate of $F$ has been obtained, its associated amplitude $A$ in \eqref{eq:M} can be recovered through taking the square root and re-normalization. These facts will be useful in the sections below. 

\section{Non-Local Means Filter}\label{Sec:NLMeans}
The concept of NLM filtering was first proposed in \cite{BuadesImage05} for the case of zero-mean AWG noise contamination. Let $X_s$ and $Y_s$ denote the intensities of the original ($X$) and observed ($Y$) image, respectively, corresponding to pixel $s \in I \equiv \{ 0, 1, \ldots, N-1 \}$, where $N$ denotes the total number of pixels in the image. Then, the filter of \cite{BuadesImage05} assumes $Y_s$ to be a realization of a mixture of ergodic processes, in which case the value of $X_s$ can be estimated by averaging $\{Y_t\}_{t \in J_s}$, where $J_s \subseteq I$ denotes the set of pixel indices whose associated intensities are distributed identically to $Y_s$. It goes without saying that, in a real-life scenario, one might not be given an oracle who would provide us with all possible sets $\{J_s\}_{s \in I}$. In this case, it makes sense to replace the binary (hard) weighting by a fuzzy (soft) one, and compute an estimate $\hat{X}_s$ of $X_s$ according to
\begin{equation}\label{OM2}
\hat{X}_s = \frac{1}{C_s} \sum_{t \in I} w_{s,t} \, Y_t, \quad \mbox{ with } C_s = \sum_{t \in I} w_{s,t},
\end{equation}
where $w_{s,t} \ge 0$  quantifies the ``contribution" of a target pixel $t \in I$ to the estimate of the source pixel $s$. Note that, ideally, the weights $w_{s,t}$ should reflect a degree of similarity between the image intensities in the vicinity of pixels $s$ and $t$ of the original image. Thus, for example, the original choice of \cite{BuadesImage05} was     
\begin{equation}\label{eq:WeightNonLocal}
w_{s,t}=\exp\left(-\frac{1}{h}\sum_{k \in \Omega}  \beta_k \, |Y_{s-k}-Y_{t-k}|^2\right),
\end{equation}
where $\Omega$ is the index set representing a symmetric neighbourhood of the centre of image coordinates. (Thus, for example, with $s$ corresponding to two spatial coordinates $s \leftrightarrow (x,y)$, $\Omega$ could be defined as $\Omega = \{ (x,y): |x| \le L_x, |y| \le L_y\}$ for some positive integers $L_x$ and $L_y$.) The ``fine-tuning" parameters $\{\beta_k\}_{k \in \Omega}$ in \eqref{eq:WeightNonLocal}  are intended to weight the domain of summation and they should be chosen to satisfy $\sum_{k \in \Omega} \beta_k = 1$, while $h > 0$ controls the overall amount of smoothing imposed by the filter. Specifically, the higher values of $h$ tend to result in overly smoothed output images, whereas smaller values produce rather milder filtering effects. As a general rule, an optimal value of $h$ should be chosen adaptively according to the level of noise in $Y$. 

To facilitate our considerations, we note that the NLM weights in \eqref{eq:WeightNonLocal} can be alternately expressed as \cite{Deledalle09, Teuber11}
\begin{equation}\label{eq:wSNL}
w_{s,t}=\prod_{k \in \Omega} ({\rm SNL}_{s,t,k})^{\frac{\beta_k}{h}},
\end{equation}
with ${\rm SNL}_{s,t,k}$ being a Gaussian SM defined as
\begin{equation}\label{eq:SimGauss}
{\rm SNL}_{s,t,k} = \exp \left(-|Y_{s-k}-Y_{t-k}|^2\right).
\end{equation}
It is important to note that the value of ${\rm SNL}_{s,t,k}$ is always bounded between 0 and 1, and it can be shown (see  \cite{Kervrann07bayesiannon-local, Deledalle09} for more details) that its choice in \eqref{eq:wSNL} and \eqref{eq:SimGauss} is optimal in the case of AWG noise contamination.

It should be noted that the use of weights $\{w_{s,t}\}$ for linearly weighting the noise samples $Y_t$ is not the only way in which the weights can be ``intermingled" with noisy data to yield an NLM estimate $\hat{X}_s$ similar to \eqref{OM2}. 
Thus, for instance, an alternative way of weighting is used in \cite{Deledalle09,Teuber11} based on the maximum likelihood (ML) framework. Apart from the  ML-based weighting scheme, the works in \cite{Deledalle09, Teuber11} suggest a unified approach to computation of the optimal weights based on the formal (and, in general, non-Gaussian) statistical properties of the original image as well as of measurement noise. Despite the generality of the above formulation, however, its application to the case of MR imagery results in SMs which possess a number of undesirable properties. In the section that follows, these properties are brought under consideration, followed by the derivation of an original methodology that allows one to fix them.

\section{Statistical approaches to computation of SM for MRI}\label{Sec:PossibleApproach}
In \cite{Deledalle09}, it was suggested to set the SM ${\rm SNL}_{s,t,k}$ to be equal to the posterior probability of $X_{s-k}=X_{t-k}$ conditioned on observations of $Y_{s-k}$ and $Y_{t-k}$. Formally,
\begin{equation}\label{eq:SNL=Posterior}
{\rm SNL}_{s,t,k}=P(X_{s-k}=X_{t-k} | Y_{s-k}, Y_{t-k}), 
\end{equation}
which, for the case $\beta_k/h = 1, \forall k \in \Omega$, leads to following definition of the NLM weights
\begin{equation}\label{OM3}
w_{s,t}= \prod_{k \in \Omega} P(X_{s-k} = X_{t - k} | Y_{s-k}, Y_{t-k}).
\end{equation}
It is worthwhile noting that, under the assumption of statistical independence of the intensities of the original image $X$, the weights in \eqref{OM3} can be viewed as the posterior probability of the image patches $\{X_{s-k}\}_{k \in \Omega}$ and $\{X_{t-k}\}_{k \in \Omega}$ to consist of the same intensities, conditioned on the  observation of their corresponding noisy values $\{Y_{s-k}\}_{k \in \Omega}$ and $\{Y_{t-k}\}_{k \in \Omega}$, respectively \cite{Deledalle09}. Although the assumption of statistical independence is an obvious oversimplification, it is often employed in NLM filtering to render the final estimation scheme computationally feasible. 

It should be noted that the SM in (\ref{eq:SNL=Posterior}) seems to have a serious theoretical flaw in the case of continuous random variables $X_{s-k}$ and $X_{t-k}$, in which case the SM is always equal to zero \cite[p. 111]{Grimmett01}. To overcome this difficulty, it was suggested in \cite{Teuber11} to introduce an auxiliary random variable $U_k \equiv X_{s-k}-X_{t-k}$ and set the SM ${\rm SNL}_{s,t,k}$ to the value of the conditional density $p_{U_k|Y_{s-k},Y_{t-k}}(u_k|y_{s-k},y_{t-k})$ at $u_k=0$. Alternatively, one can use a different auxiliary variable $V_k=X_{s-k}/X_{t-k}$, in which case the SM can be set to be equal to the value of $p_{V_k|Y_{s,k},Y_{t,k}}(v_k|y_{s-k},y_{t-k})$ at $v_k=1$ \cite{Teuber11}. For the convenience of referencing, the above SMs will be referred below to as the {\em subtractive} and the {\em rational} SMs, respectively. Note that, although similar in their underlying philosophy, these measures lead to substantially different de-noising schemes, as it is detailed below.

In the present work, we explore both the subtractive and rational SMs for two different types of input data, namely for the measured magnitude $M$ and its squared normalized version $G$. The main contribution of the next propositions is to provide closed-form expressions for the SMs which result from using a subtractive $U_k$ on $G$ images, and a rational $V_k$ on $M$ images. Unfortunately, for the remaining two combinations ({\em viz.}, ``rational" $G$ and ``subtractive" $M$) it does not seem to be possible to derive closed-form expressions as well. In these cases, the measures need to be computed numerically -- the approach which should be avoided in practice due to its low computational efficiency.       

\begin{theorem}\label{Prop:SubtractiveNCCS}
Let $G = (M/\sigma)^2$ be a squared and normalized version of the MR magnitude image $M$ as given by \eqref{eq:M}. Moreover, let $F = (A/\sigma)^2$, where $A$ denotes the true signal amplitude. Then, the subtractive SM $\MS$ is given by
\begin{equation}\label{eq:SubtractiveNCCS}
\MS = p_{F_{s-k}-F_{t-k} \mid G_{s-k}, G_{t-k}}(0 \mid g_{s-k}, g_{t-k})=\frac{1}{4} \, e^{-(g_{s-k}+g_{t-k})/4} I_0 \left(\frac{\sqrt{g_{s-k} \, g_{t-k}}}{2}\right).
\end{equation}
\end{theorem}

The proof of Proposition \ref{Prop:SubtractiveNCCS} is provided in Appendix \ref{App:NCCS}, while Fig.~\ref{Fig:SNCCS} shows a number of $\MS$ curves corresponding to various values of $g_{t-k}$ and some fixed values of $g_{s-k}$.

\begin{figure}[ht]
\centering
\subfigure[]{
   \includegraphics[scale=0.3]{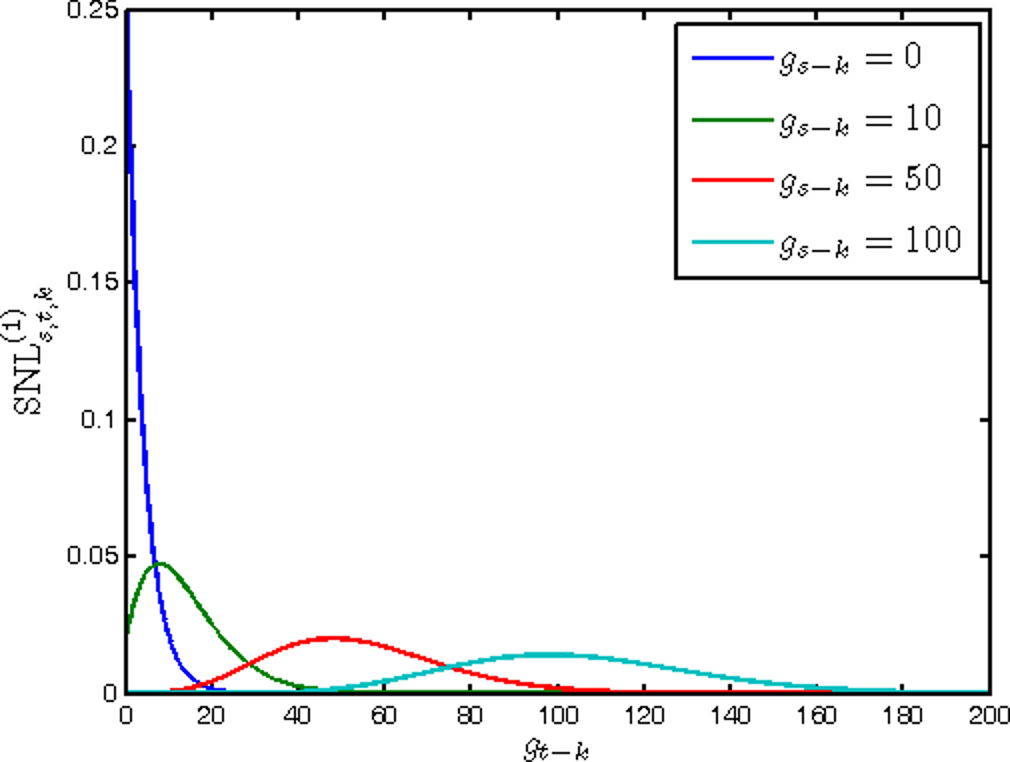}
   \label{Fig:SNCCS}
 }
 \subfigure[]{
   \includegraphics[scale=0.3]{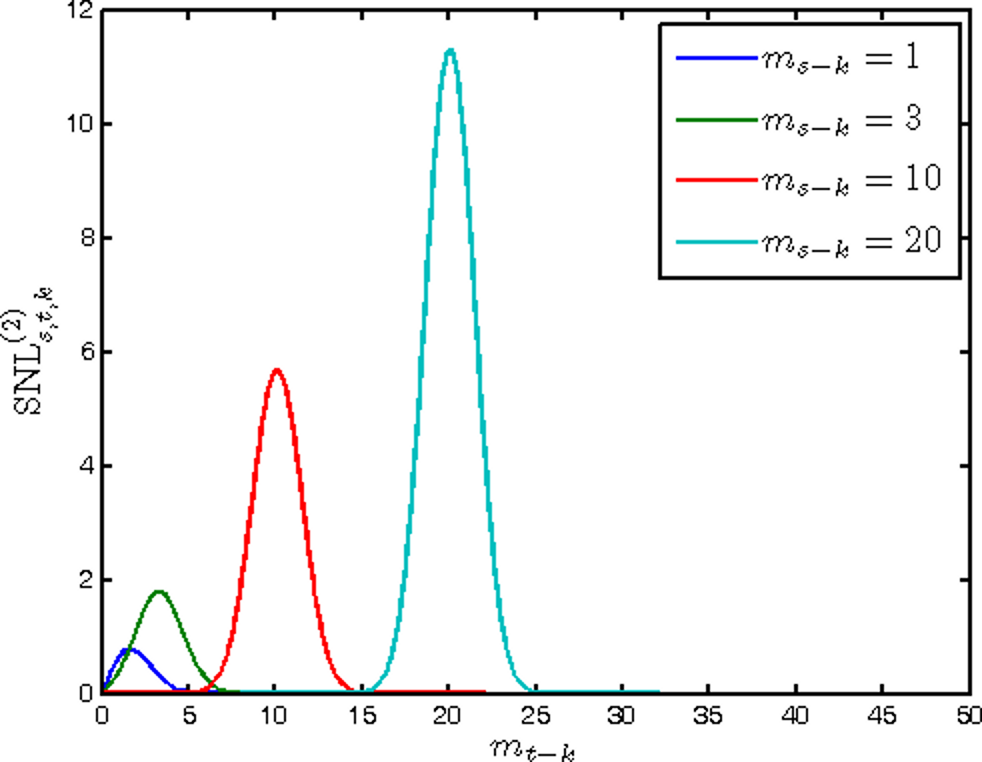}
   \label{Fig:RatioRice}
 }
\label{fig:SNL}
\caption{(a) Subtractive similarity measure $\MS$; (b) Rational similarity measure $\MR$.}
\end{figure}

Observing Fig.~\ref{Fig:SNCCS}, a number of critical remarks are in order.  

\begin{enumerate}
\item The smaller the value of $g_{s-k}$ is, the narrower is the effective support of $\MS$. This effect is due to the signal-dependent Gaussian noise $2 \sqrt{F} \xi$ in \eqref{eq:G}. Indeed, smaller values of $g_{s-k}$ suggest smaller values of their related $f_{s-k}$, and therefore smaller values of the above mentioned Gaussian noise component. In such a case, $\MS$ becomes more sensitive to the value of $g_{s-k}-g_{t-k}$ (fast decay), which indicates that the difference $f_{s-k}-f_{t-k}$ is likely to be small as well. On the other hand, the contribution of the (signal-dependent) Gaussian noise component becomes stronger for relatively large values of $g_{s-k}$ (and hence of $f_{s-k}$). In this case, $\MS$ has a slower convergence rate, since larger values of $g_{s-k} - g_{t-k}$ no longer imply a larger discrepancy between $f_{s-k}$ and $f_{t-k}$. 
\item The smaller the value of $g_{s-k}$ is, the more heavy-tailed is the behaviour of $\MS$. This effect can be attributed to the dominance of the exponential noise component $\eta$ in (\ref{eq:G}). Moreover, when $g_{s-k}$ is large (which implies the dominance of the Gaussian noise), the $\MS$ curves appear to be more symmetric.
\end{enumerate}

The proposition that follows extends the previous results to the case of magnitude signals and a rational SM.
\begin{theorem}\label{Prop:RatioRice}
Let $M$ be the MR magnitude image as given by \eqref{eq:M}. Moreover, let $A$ be the original signal amplitude. Then, the rational SM $\MR$ is given by
\begin{equation}\label{eq:RatioRice}
\MR = p_{A_{s - k}/A_{t - k} \mid M_{s-k}, M_{t-k}}(1 \mid m_{s-k}, m_{t-k}) =
\frac{m_{s-k} \, m_{t-k}}{2\sigma^2} \, e^{-(m_{s-k}^2+m_{t-k}^2)/4\sigma^2} I_0\left(\frac{m_{s-k} \, m_{t-k}}{2\sigma^2}\right).
\end{equation}
\end{theorem}
The proof of Proposition \ref{Prop:RatioRice} can be found in Appendix \ref{App:Rician}. The plot of $\MR$ is shown in Fig. \ref{Fig:RatioRice}, where each curve is drawn with a fixed $m_{s-k}$ and varying $m_{t-k}$, while $\sigma$ is set to be equal to 1.

The main observations in the above case are:
\begin{enumerate}
\item As the value of $m_{s-k}$ increases, the noise distribution tends towards Gaussian, and as a result, the similarity measure $\MR$ becomes more symmetric and appears to have the shape of a Gaussian SM.
\item For relatively small values of $m_{s-k}$, on the other hand, the shape of $\MR$ is noticeably asymmetric, being heavy-tailed towards the right.  
\end{enumerate}

Although $\MS$ and $\MR$ appear to reflect the main properties of their corresponding noise distributions, they share a number of critical drawbacks which we point out below. 
\begin{enumerate}
\item Neither $\MS$ nor $\MR$ (as given by \eqref{eq:SubtractiveNCCS} and \eqref{eq:RatioRice}, respectively) attain their maxima values at the point when the two arguments received by the measures are equal. Particularly, for a fixed value of $g_{s-k}$ (resp. $m_{s-k}$), the $\MS$ measure (resp. the $\MR$ measure) is maximal at some $g_{t-k} < g_{s-k}$  (resp. $m_{t-k} < m_{s-k}$).
\item The maximal values $\MS$ and $\MR$ can attain depend on the values of the ``source" intensities $g_{s-k}$ and $m_{s-k}$, respectively. In other words, the value of $\MS (\alpha, \alpha)$ (resp. $\MR (\beta, \beta)$) depends on the value of $\alpha$ (resp. $\beta$). From a purely applicational point of view, this fact suggests that the measures are not scale invariant, and as a result, the weights $w_{s,t}$ in \eqref{OM2} are defined {\em not only} by how dissimilar compared intensities are, but also by their absolute values.
\item As can be seen from Figure~\ref{Fig:RatioRice}, $\MR$ can become unbounded which is not a favourable property of the similarity measure. 
\end{enumerate}

Some of the above mentioned limitations of $\MS$ and $\MR$ have already been pointed out in \cite{Teuber11}, where the authors apply the method of \cite{Deledalle09} to multiplicative noise. The main conclusion which one can immediately draw from the discussion above as well as based on \cite{Teuber11} is that neither $\MR$ nor $\MS$ is optimal to deal with the cases of Rician and/or NCCS noises, which are the most relevant types of noises in MRI. In the next section, we propose a new SM, which is free of the limitations of $\MS$ and $\MR$ mentioned above.

\section{Proposed Approach}\label{Sec:Proposed}
\subsection{Subtractive SM for NCCS noise}
To derive the proposed SM in a consistent and intuitive way, we start with the case of the subtractive SM (SSM), which is defined as \cite{Teuber11}
\begin{align}\label{eq:AddMeasure}
{\rm SSM} &= p_{F_{s-k}-F_{t-k} \mid G_{s-k}, G_{t-k}}(0 \mid g_{s-k}, g_{t-k}) = \notag \\
&= \int_{0}^\infty p_{F_{s-k} \mid G_{s-k},G_{t-k}}(f \mid g_{s-k}, g_{t-k}) \, p_{F_{t-k} \mid G_{s-k},G_{t-k}}(f|g_{s-k},g_{t-k}) \, df.
\end{align}
For the case of a NCCS distribution, one alternatively has (see Appendix \ref{App:NCCS} for the details of derivation)
\begin{align}\label{OM4}
{\rm SSM} &\equiv \MS = p_{F_{s-k} - F_{t-k} \mid G_{s-k}, G_{t-k}} ( 0 \mid g_{s-k}, g_{t-k}) = \notag \\
&= \int_{0}^\infty p_{G_{s-k} \mid F_{s-k}}(g_{s-k} \mid f) \, p_{G_{t-k} \mid F_{t-k}} (g_{t-k} \mid f) \, df.
\end{align}
In the above equation, it has been assumed that the prior probability $p_{F_{z-k}}$ (where $z \in \{s, t\}$ and $k \in \Omega$) is uniform. Moreover, it has also been assumed that, given the noisy intensity value at a particular location, its original (i.e., noise-free) value is conditionally independent of noisy intensities at different locations ({\it viz.}, $p_{F_{z_1-k} \mid G_{z_1-k}, G_{z_2-k}} =  p_{F_{z_1-k} \mid G_{z_1-k}}$). Finally, it also deserves noting that, for a general noise distribution, it can be shown that the relation in \eqref{OM4} holds with the {\em proportionality} rather than the equality sign \cite{Deledalle09, Teuber11}. Specifically,  
\begin{equation}\label{eq:AdditiveProp}
{\rm SSM} = p_{F_{s-k}-F_{t-k} \mid G_{s-k}, G_{t-k}} (0 \mid g_{s-k}, g_{t-k}) \propto \int_{0}^\infty
p_{G_{s-k} \mid F_{s-k}}(g_{s-k} \mid f) \, p_{G_{t-k} \mid F_{t-k}}(g_{t-k} \mid f) \, df.\end{equation}

To overcome the limitations of $\MS$ and $\MR$ as per the discussion in the previous section, we interpret the right-hand side of \eqref{eq:AdditiveProp} as an inner product between the {\em likelihood functions} $L_{g_{s-k}}(\cdot)$ and $L_{g_{t-k}}(\cdot)$, with $L_{g}(f)$ given by
\begin{equation}
L_{g}(f)=p_{G \mid F}(g \mid f).
\end{equation}
In other words, the SM derived from the probabilistic point of view has the form of the inner product of likelihood functions, where each of the likelihood functions is indexed by its corresponding noisy observation. Note that, in general, the likelihood functions $L_{g_{s-k}}(\cdot)$ and $L_{g_{t-k}}(\cdot)$ have unequal norms, and as a result their inner product is not maximized when $g_{s-k}=g_{t-k}$ (which would be a natural and desirable property of an SM to have). To overcome this shortcoming, we suggest to normalize the inner product, thereby converting it into a correlation similarity measure (CSM) according to
\begin{equation}\label{eq:CorrSim} 
{\rm CSM}_{s,t,k}^{(1)} = \frac{ \langle L_{g_{s-k}}, L_{g_{t-k}} \rangle}{ \| L_{g_{s-k}} \|_2 \, \|L_{g_{t-k}}\|_2},
\end{equation}
where $\langle x, y \rangle = \int_{0}^\infty x(f) y(f) \, df$ and $\|x\|_2 = \sqrt{\langle x, x\rangle}$. It is interesting to observe that ${\rm CSM}_{s,t,k}^{(1)}$ can be also viewed as an inner product of two functions lying on the unit sphere in $\mathbb{L}_2(\mathbb{R}^+)$. Therefore, ${\rm CSM}_{s,t,k}$ has an interpretation of the cosine of the angle between the two functions.    

The CSM in \eqref{eq:CorrSim} turns out to be particularly advantageous in the case of MR imagery. First, it is free of all the major limitations of $\MS$ and $\MR$ as previously discussed. In particular, ${\rm CSM}_{s,t,k}$ is always smaller or equal to 1, and it achieves its maximum value when $g_{s-k}=g_{t-k}$. Secondly, for the case of NCCS noises, the CSM measure can be shown to  have a neat closed-form expression which is given by
\begin{equation}\label{eq:CorrNCCS}
\CSMS = \frac{I_0\left( \sqrt{\hat{g}_{s-k} \, \hat{g}_{t-k}} \right)}{\sqrt{I_0(\hat{g}_{s-k}) \, I_0(\hat{g}_{t-k})}}, \,\, \mbox{ where } \hat{g} =  g/2.
\end{equation}

\begin{figure}[ht]
\centering
\subfigure[]{
   \includegraphics[scale=0.3]{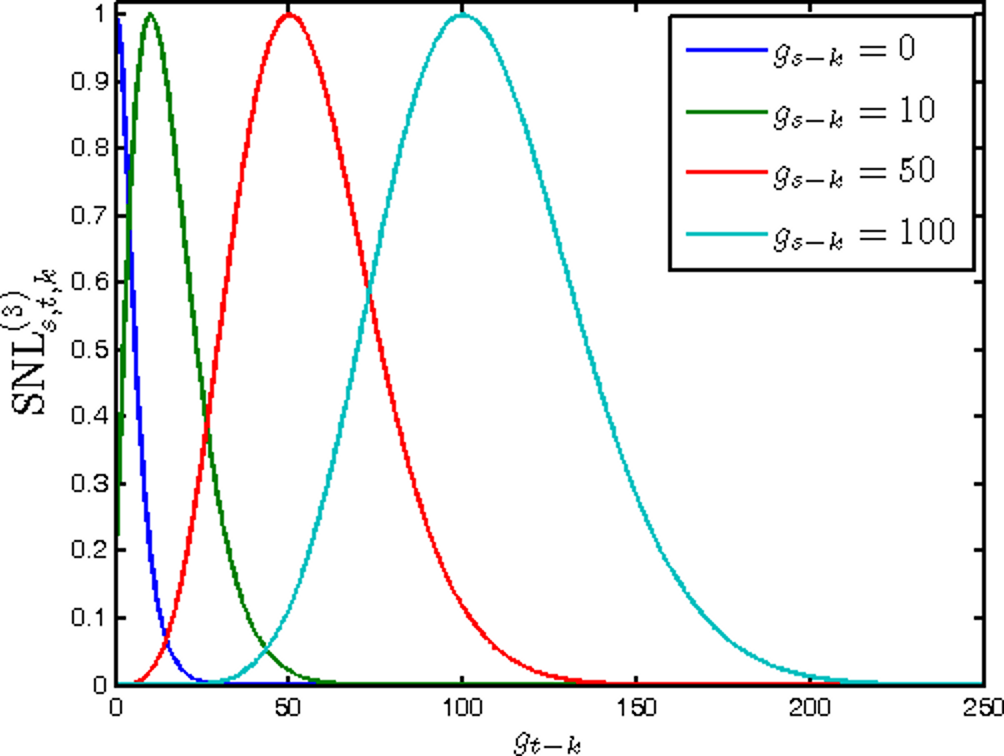}
   \label{Fig:CorrNCCS}
 }
 \subfigure[]{
   \includegraphics[scale=0.3]{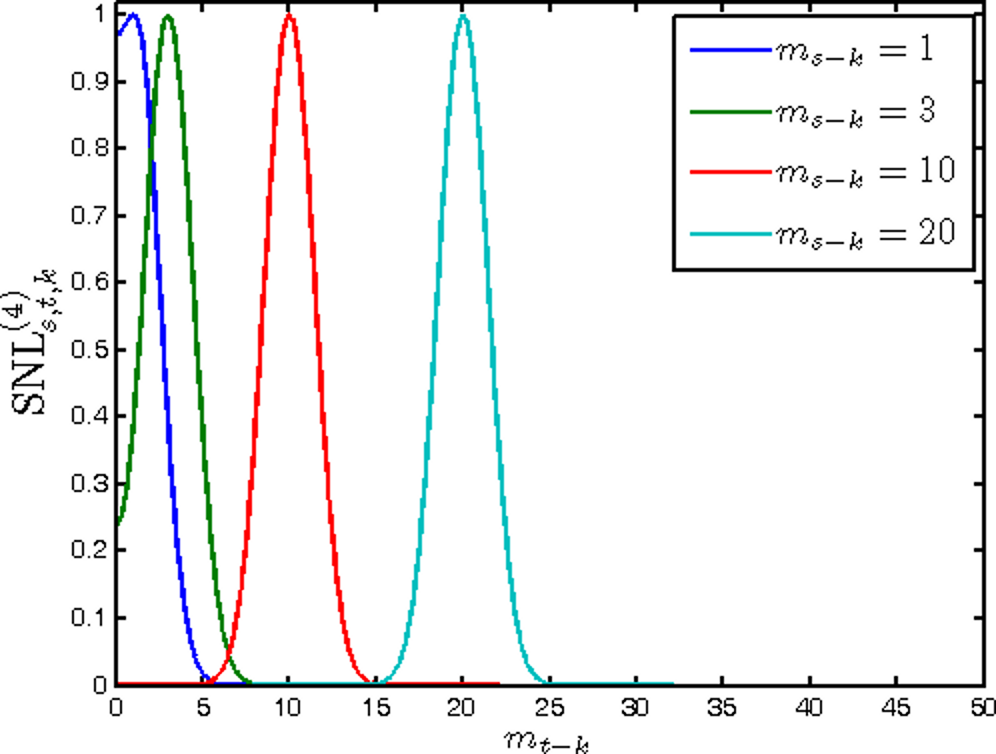}
   \label{Fig:CorrRice}
 }
\label{fig:CorrSNL}
\caption{(a) Proposed CSM $\CSMS$ for the case of NCCS noise distribution; (b) Proposed CSM $\CSMR$ for the case of Rician noise distribution.}
\end{figure}

A number of graphs of $\CSMS$ are shown in Fig. \ref{Fig:CorrNCCS}, where each curve is drawn with a fixed $g_{s-k}$ and varying $g_{t-k}$. It can be seen from the graphs that the shape of each curve is similar to those in Fig. \ref{Fig:SNCCS}. However, unlike the plots in Fig. \ref{Fig:SNCCS}, each curve in Fig. \ref{Fig:CorrNCCS} is maximized when $g_{t-k}=g_{s-k}$ and the maximum value is the same for all the curves and is equal to 1.

\subsection{Rational SM for Rician noise}
To derive an expression for a CSM for the case of Rician noise, we first recall that the rational SM (RSM) is given by \cite{Teuber11}
\begin{align}\label{OM5}
{\rm RSM} &= p_{A_{s-k}/A_{t-k} \mid M_{s-k}, M_{t-k}}(1 \mid m_{s-k}, m_{t-k}) \notag\\
&= \int_{0}^\infty a \, p_{A_{s-k} \mid M_{s-k}, M_{t-k}} (a \mid m_{s-k}, m_{t-k}) \, p_{A_{t-k} \mid M_{s-k},M_{t-k}} (a \mid m_{s-k},m_{t-k}) \, da.
\end{align}
Using a method similar to the one discussed in the previous subsection, one can obtain
\begin{align}\label{OM51}
{\rm RSM} =  \int_{0}^\infty a \, p_{M_{s-k} \mid A_{s-k}} (m_{s-k}\mid a) \, p_{M_{t-k}\mid A_{t-k}} (m_{t-k}\mid a) \, da,
\end{align}
where the equality sign can be replaced by a proportionality sign for general noise distributions, as it was done in the previous subsection. Moreover, similar to the case with $\CSMS$, the integral in \eqref{OM51} can be interpreted as a weighted inner product $\langle x, y \rangle_a = \int_{0}^\infty x(a) y(a) \, a \, da\, $, where $a \, da$ can be viewed as a  ``modified" integration measure. Using this notation, we have
\begin{equation}\label{OM6}
{\rm RSM} \equiv \left\langle L_{m_{s-k}}, L_{m_{t-k}} \right\rangle_a.
\end{equation}

As the final step, one can normalize the inner product in \eqref{OM6} in a way similar to 
\eqref{eq:CorrSim}, which leads to a different CSM for Rician noises, which is given by 
\begin{equation}\label{eq:CorrRice}
\CSMR=\frac{I_0\left(\frac{m_{s-k}m_{t-k}}{2\sigma^2}\right)}{\sqrt{I_0\left(\frac{m_{s-k}^2}{2\sigma^2}\right)I_0\left(\frac{m_{t-k}^2}{2\sigma^2}\right)}}
\end{equation}

A number of $\CSMR$ curves are shown in Fig. \ref{Fig:CorrRice} for different (fixed) values of $m_{s-k}$ and a range of varying $m_{t-k}$. Once again, one can observe that the curves are similar in shape as those of Fig. \ref{Fig:RatioRice}. However, unlike Fig. \ref{Fig:RatioRice}, each $\CSMR$ curve is maximized when $m_{t-k}=m_{s-k}$ and the maximum value is equal to 1 in each case.

There are two important facts about $\CSMS$ and $\CSMR$ that deserve to be paid special attention. In particular: 
\begin{enumerate}
\item The values of $\CSMS$ and $\CSMR$ (as given by (\ref{eq:CorrNCCS}) and (\ref{eq:CorrRice}), respectively) are equal under the substitution $g_{z-k}=(m_{z-k}/\sigma)^2$. In other words, the value of $\CSMS$ for some arbitrary $g_{s,k}$ and $g_{t,k}$ will be equal to the value of $\CSMR$ for $m_{s,k}$ and $m_{t,k}$, whenever $g_{s,k} = (m_{s,k}/\sigma)^2$ and $g_{t,k} = (m_{t,k}/\sigma)^2$, which is precisely the case at hand. In other words, the proposed SMs are equivalent under reparametrization $G = (M/\sigma)^2$. This fact suggests that, in terms of $\CSMS$ and $\CSMR$ values, two patches of an original MR image and their corresponding squared and normalized versions are equally similar.

\item Using the fact that, for sufficiently large $x$, it holds that
\begin{equation}\label{OM7}
I_0(x) \approx \frac{\exp(x)}{\sqrt{2\pi x}},
\end{equation}
and plugging \eqref{OM7} into (\ref{eq:CorrRice}) instead of the original Bessel functions, we obtain 
\begin{align}\label{eq:Optexp}
\CSMR \approx \exp\left\{-\frac{| m_{s-k}-m_{t-k} |^2}{4\sigma^2}\right\}.
\end{align}
The above approximation holds with a high precision for relatively large values of SNR (i.e., for 
$\frac{m_k}{\sigma},\frac{m_l}{\sigma}\gg 1$). This is an exceptional property of $\CSMR$, since it is known that Rician noise in MRI converges to Gaussian noise, when SNR increases. In such a case, the proposed $\CSMR$ measure converges to the form of \eqref{eq:SimGauss}, whose optimality for Gaussian noises was proven in \cite{Deledalle09}.
\end{enumerate}

\section{Bias Removal}\label{Sec:BiasRemoval}
The original NLM filter of \cite{BuadesImage05} employs the adaptive local averaging scheme of \eqref{OM2}, in which the output intensity is computed as a weighted linear combination of the measured intensities, while the weights are determined adaptively, based on an SM in use. Unfortunately, the filter in \eqref{OM2} has some optimality properties in the case of additive Gaussian noises alone, while the efficiency of the filter is known to deteriorate for some more general types of measurement noise. In the latter case, one can adopt the more general filtering procedure that was used in  \cite{Deledalle09,Teuber11}, according to which an estimated value $\hat{X}_s$ of some unknown $X_s$ is given by 
\begin{equation}\label{eq:LWMLE}
\hat{X}_s = \arg\max_{X_s} \sum_{t \in I} w_{s,t} \, \log\,p(Y_t \mid X_s),
\end{equation}
where $Y_t$ denotes the measured intensities and $\{w_{s,t}\}$ are some predefined weights. The estimate in \eqref{eq:LWMLE} has been motivated by the work in \cite{polzehl2006propagation} where it has been referred to as a weighted maximum likelihood (WML) estimate. 

Unfortunately, in the case when $Y_t$ in \eqref{eq:LWMLE} follows either a Rician or an NCCS distribution, \eqref{eq:LWMLE} does not seem to have a close-form analytical form, which necessitates the use of numerical optimization. Alternatively, one can take advantage of the relation in \eqref{eq:Eg} to estimate $\mathcal{E}\{G_s\}$ as
\begin{equation}
\mathcal{E}\{G_s\} \approx \sum_{t \in I}\, w_{s,t} \, g_t, 
\end{equation}
followed by estimating the related original amplitude $A_s = \sigma \sqrt{\mathcal{E}\{G_s\} - 2}$ as
\begin{equation}\label{eq:Ahat}
A_s \approx \sigma \bigg[ \max \Big\{ \Big[ \sum_{t \in I} w_{s,t} \, g_t \Big] - 2, \, 0 \Big\} \bigg]^{1/2} \equiv \hat{A}^{(1)}_s.
\end{equation}
where the $\max(\cdot)$ operator is used to avoid complex estimates. Note that, even though applying the $\max(\cdot)$ operator may seem like a purely ad-hoc procedure, it has been shown to be optimal in the ML sense in \cite{Sijbers98}. This estimate has also been used previously in \cite{Daessle08} in the context of non-local means.

Yet another option to estimate $A_s$ is to apply averaging to $M$ (as opposed to $G$). Although the square of the average value is, in general, not equal to the average of the squared values, a common method of estimating ${A}_s$ is \cite{Manjon08}
\begin{equation}\label{eq:AsM}
A_s \approx \bigg[ \max \Big\{ \Big[ \sum_{t \in I} w_{s,t} \, m_t \Big]^2 - 2 \sigma^2, \, 0 \Big\} \bigg]^{1/2} \equiv \hat{A}^{(2)}_s.
\end{equation}

\begin{figure}[h!]
\centering
\subfigure[]{
   \includegraphics[scale =0.25] {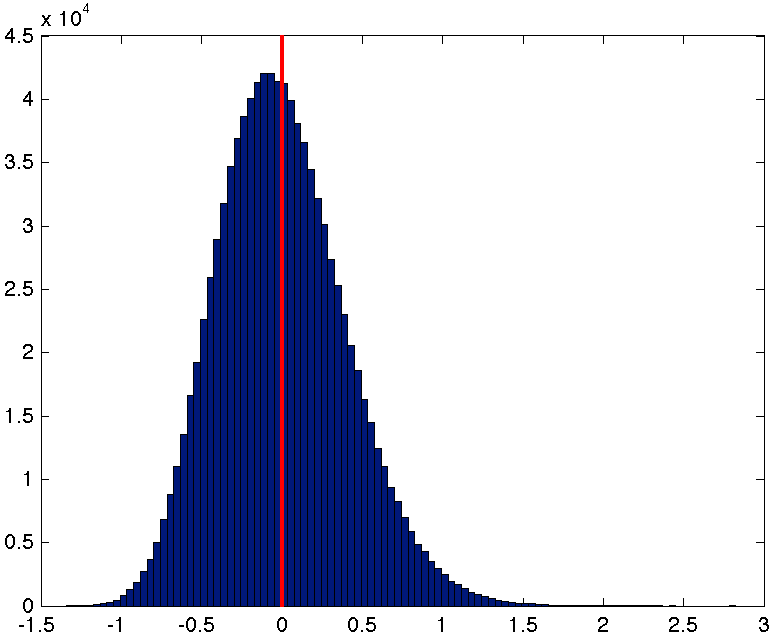}
   \label{fig:hist1}
 }
 \subfigure[]{
   \includegraphics[scale =0.25] {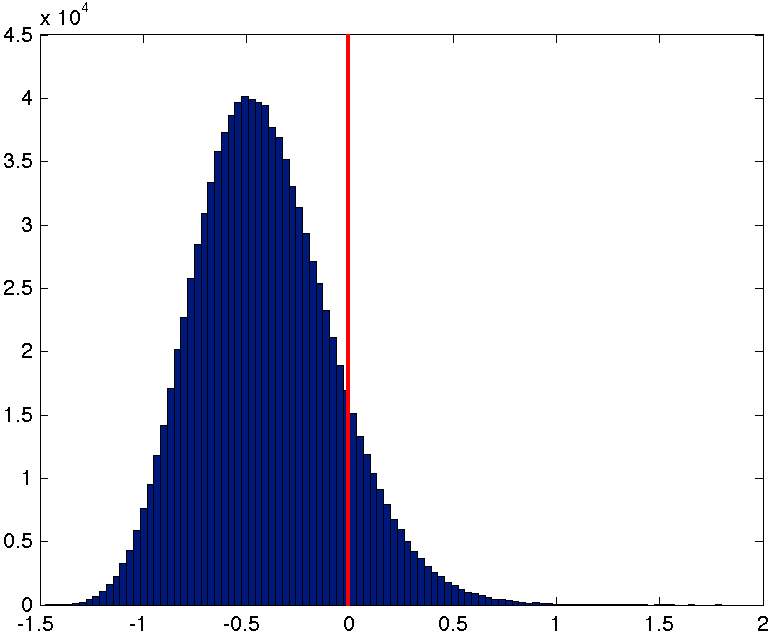}
   \label{fig:hist2}
 }
\caption{(a) Histogram of $(\hat{A}^{(1)}_s)^2$; (b) Histogram of $(\hat{A}^{(2)}_s)^2$.}
\label{fig:Histograms}
\end{figure}

Despite the fact that $\left[ \sum_{t \in I} w_{s,t} \, m_t \right]^2 / \sigma^2$ does not seem to be a legitimate estimate of $\mathcal{E}\{G_s\}$, the estimate $\hat{A}^{(2)}_s$ in \eqref{eq:AsM} often produce more accurate reconstruction results in terms of the mean-squared error (MSE) as compared to the estimate $\hat{A}^{(1)}_s$ in \eqref{eq:Ahat}. To understand the reasons which underpin the above phenomenon, it is instructive to consider the following numerical experiment. Fig.~\ref{fig:hist1} and Fig.~\ref{fig:hist2} show the histograms of the estimates in \eqref{eq:Ahat} and \eqref{eq:AsM}, respectively, which have been computed {\em without} applying the square root and the $\max(\cdot)$ operator. With a slight abuse of notations, these estimates are referred to below as $(\hat{A}^{(1)}_s)^2$ and $(\hat{A}^{(2)}_s)^2$ (which, despite the square in their superscripts, are allowed to have an arbitrary sign). In both cases, the original amplitude $A$ was set to be equal to zero (i.e. $A = 0$), the noise variance $\sigma$ was normalized to have the value of $\sigma = 1$, the weights $w_{s,t}$ were chosen to correspond to uniformly averaging 25 random realizations of $g_t$ and $m_t$, respectively\footnote{In practice, 25 appears to be a typical size of the neighbourhood $I$ used for NLM filtering.}, and both histograms have been computed based on the results of $10^6$ independent trials. Observing Fig.~\ref{fig:Histograms} one can see that both estimates provide outputs in the vicinity of the true squared amplitude $A^2=0$. However, while the histogram of $(\hat{A}_s^{(1)})^2$ is shaped more or less symmetrically around the origin, the histogram of $(\hat{A}_s^{(2)})^2$ is noticeably biased to the left, which suggests that the estimate in \eqref{eq:AsM} tends to {\em underestimate} the true squared magnitude of $A^2=0$. In terms of the probabilities $P((\hat{A}_s^{(1)})^2 \le 0)$ and $P((\hat{A}_s^{(2)})^2 \le 0)$ with which the estimates yield non-positive outputs, one can therefore conclude that $P((\hat{A}_s^{(1)})^2 \le 0) < P((\hat{A}_s^{(2)})^2 \le 0)$. (Thus, for example, in the case of Fig.~\ref{fig:Histograms}, $P((\hat{A}_s^{(1)})^2 \le 0) \approx 0.54$, while $P((\hat{A}_s^{(2)})^2 \le 0) \approx 0.89$.) This fact, in turn, implies that, after applying the $\max(\cdot)$ operator, the estimation in \eqref{eq:AsM} is much more likely to produce the exact reconstruction of $A=0$ as compared to the estimation in \eqref{eq:Ahat}.
 
In MRI, zero-valued amplitudes $A$ are predominant at the background areas of MR images, which are normally devoid of water content. At such areas, therefore, the estimation in \eqref{eq:Ahat} should be expected to produce larger values of MSE as compared to the case of \eqref{eq:AsM}. We will have more to add to the subject in the following sections of the paper.

\section{Results}\label{Sec:Results}
\subsection{Reference Methods}
The performance of the proposed method has been compared with several standard and established algorithms. Specifically, as the first reference method, the total-variation filter of \cite{Rudin:1992fh}, implemented by means of the fast fixed-point algorithm of \cite{Chambolle:2004uq}, has been used. The algorithm (which is referred to below as TVDN, an acronym of total variation denoising) has been applied to the magnitude MR data, followed by the bias-removal procedure specified by \eqref{eq:AsM}. As the second reference method, the wavelet-based method of \cite{Pizurica03} has been used\footnote{The method was implemented using the code available at the author's webpage at {\tt http://telin.ugent.be/\~{}sanja/Sanja\_files/} {\tt Software/MRIprogram.zip.}}. In what follows, this method is referred to as wavelet de-noising (WDN). As the final method used for numerical comparison, the NLM filter of \cite{Manjon08} has been employed. Since the filter uses Gaussian weights to compute the similarity measure, we refer to this method as GNLM, an acronym for Gaussian NLM. In the case of all reference methods under comparison, their respective parameters have been set based on the guidelines specified in their associated papers.

Two new approaches to NLM filtering of MR images are proposed in this paper. Specifically, the first approach is applied to squared-magnitude MR data using the subtractive CSM of \eqref{eq:CorrNCCS}, followed by the bias-correction procedure given in \eqref{eq:Ahat}. For the convenience of referencing, this filtering approach is referred below to as NLMS (with ``S" standing for ``subtractive"). The second approach, on the other hand, is applied to magnitude data using the rational CSM of \eqref{eq:CorrRice}, followed by the bias-correction procedure of \eqref{eq:AsM}. In what follows, this filtering approach is referred to as NLMR (with ``R" standing for ``rational"). All the acronyms of the proposed and reference algorithms are summarized in Table \ref{Table:2}.

\begin{table}
\centering
\caption{Acronyms of the proposed and reference algorithms}
\label{Table:2}
\begin{tabular}{c | c | c}
\hline
Algorithm name  & Reference & Input image type\\
\hline
\hline
TVDN 	& 	\cite{Rudin:1992fh}	& M\\
WDN 	& 	\cite{Pizurica03} 	& G\\
GNLM 	& 	\cite{Manjon08} 	& M\\
NLMS 	& 	Proposed 			& G\\
NLMR 	& 	Proposed 			& M\\
\hline
\end{tabular}
\end{table}

\subsection{Simulated Data}
\begin{figure}[h!]
\centering
\subfigure[]{
   \includegraphics[scale =0.65] {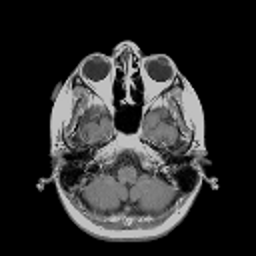}
   \label{fig:Sim1Orig}
 }
 \subfigure[]{
   \includegraphics[scale =0.65] {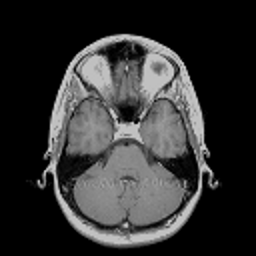}
   \label{fig:Sim2Orig}
 }
\subfigure[]{
   \includegraphics[scale =0.65] {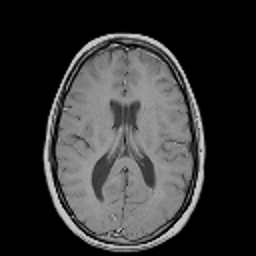}
   \label{fig:Sim4Orig}
 }
\caption{(a) Test slice \#4, (b) Test slice \#7 and (c) Test slice \#16 of the MATLAB\textsuperscript{\textregistered} MRI database.}
\label{fig:SimOrig}
\end{figure}

In this subsection, the MRI images from the built-in MRI dataset available in the MATLAB\textsuperscript{\textregistered} toolbox have been used as test subjects. Specifically, the denosing algorithms have been tested using the axial slices number 4, 7 and 16 (shown in Figures \ref{fig:Sim1Orig}, \ref{fig:Sim2Orig} and \ref{fig:Sim4Orig}, respectively), which represent a spectrum of different cerebral structures. For quantitative comparison, simulated data have been obtained by subjecting the original test images to various levels of Rician noise. 

\subsubsection{Data Generation}
For the current test setup, given a noise-free intensity $A_k$ of the original image $A$, its noisy counterpart $M_k$ can be simulated according to   
\begin{equation}
M_k=\sqrt{(A_k+n_r)^2+n_i^2}
\end{equation}
where $n_r, \, n_i\sim \mathcal{N}(0,\sigma^2)$ are independent Gaussian random variables, which are also assumed to be independent across the image domain. In our simulation study, the standard deviation $\sigma$ was set to 10 percent of the maximum value of the original images, thereby resulting in a (peak) signal-to-noise ratio of ${\rm SNR} = A_{max}/\sigma = 10$.

\subsubsection{Performance Metrics}\label{PerformanceMertics}
Evaluating the quality of denoising in medical imaging is of subjective nature, as it is often based on the particular requirements of a medical expert. For this reason, in many cases a noisy version of an image is given preference over a de-noised version, as denoising procedures have an inherent risk of removing small structures from images. Hence, it is difficult to define an objective criterion for evaluating the quality of medical images. Nevertheless, there is a number of standard evaluation metrics used in the literature, some of which we adopt in the present study. Specifically, one of such metrics is the {\em root mean square error} (RMSE), which can be expressed in dB as given by
\begin{equation}\label{OM8}
\mbox{RMSE} = 20 \log_{10} \left[ \frac{1}{N} \sum_{k=1}^N |e_k|^2 \right]^{1/2}, 
\end{equation}
where $e_k=A_k-\hat{A}_k$ denotes the difference between the original intensity $A_k$ and its estimated value ${\hat A}_k$ at position $k$, and $N$ stands for the total number of image pixels.

When estimate ${\hat A}_k$ is biased, the mean value of $e_k$ may not be equal to zero, in general. In this case, it makes sense to replace $e_k$ in \eqref{OM8} by its centred version $e_k - \bar{e}$, with $\bar{e}$ being the sample mean of $e_k$ given by 
\begin{equation}
\bar{e}=\frac{1}{N}\sum_{k=1}^N e_k.
\end{equation}
The resulting metric is called the {\em centred} RMSE (cRMSE), and it is formally defined as
\begin{equation}\label{OM9}
\mbox{cRMSE} = 20 \log_{10}\left[ \frac{1}{N} \sum_{k=1}^N |e_k-\bar{e}|^2 \right]^{1/2}.
\end{equation}

It should be noted that, while structurally similar, the RMSE and cRMSE metrics provide different quantitative assessment in the case of biased estimation. Consequently, the analysis and comparison of both these metrics can be helpful in evaluating the performance of the de-biasing procedures detailed in Section~\ref{Sec:BiasRemoval}.

Despite the relatively straightforward interpretation offered by the RMSE and cRMSE metrics in \eqref{OM8} and \eqref{OM9}, it has been argued in \cite{Wang04} that these metrics may not always adequately represent the effect/size of certain estimation artifacts/noises. As an alternative, a different performance metric -- called the {\em structural similarity index} (SSIM) -- has been proposed. This metric has also been used in our comparative study. 

\subsubsection{Details on the choice of parameters}
In order to facilitate the reproducibility of the proposed algorithms and their results, some principal details on the choice of algorithms' parameters are specified next. In particular, all the NLM algorithms have been implemented using a square $5\times 5$ neighbourhood $\Omega$ and an $11\times 11$ search window. (Note that the latter suggests that the averaging in \eqref{OM2} was carried out only over pixels $t$ which were located within an $11\times 11$ window centred at pixel $s$.) The weights $\beta_k$ in \eqref{eq:wSNL} were defined to correspond to a separable binomial mask of the third order\footnote{In practice, this mask can be computed as $v v^T$, with $v = 2^{-4}\cdot [1~4~6~4~1]^T$.}, while the tuning parameter $h$ was set to be equal to 0.4. Note that, in general, it has been found empirically that the proposed algorithms performed the best for $h \in [1/3, \,1/2]$. 

\subsubsection{Comparative analysis of algorithm performance}
\begin{figure}[h!]
\centering
\subfigure[]{
   \includegraphics[scale =0.65] {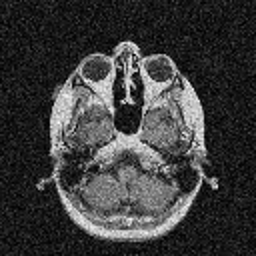}
   \label{fig:Sim1Noisy}
 }
 \subfigure[]{
   \includegraphics[scale =0.65] {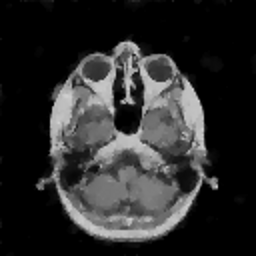}
   \label{fig:Sim1TV}
 }
\subfigure[]{
   \includegraphics[scale =0.65] {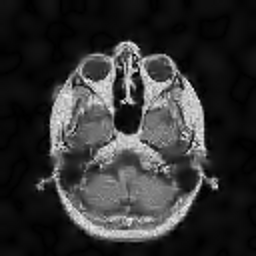}
   \label{fig:Sim1Pizurica}
 }
 \subfigure[]{
   \includegraphics[scale =0.65] {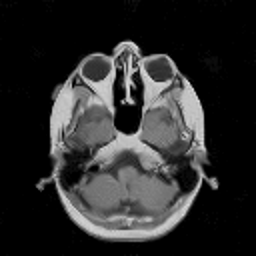}
   \label{fig:Sim1Manjon}
 }
 \subfigure[]{
   \includegraphics[scale =0.65] {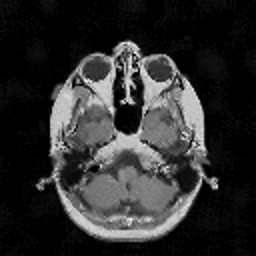}
   \label{fig:Sim1NCCS}
 }
 \subfigure[]{
   \includegraphics[scale =0.65] {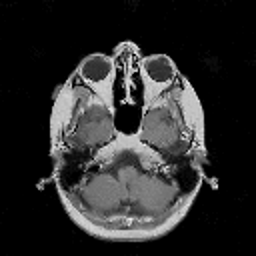}
   \label{fig:Sim1Rice}
 }
\caption{(a) Noisy version of slice \#4; (b)-(f) Reconstruction results obtained using TVDN, WDN, GNLM, NLMS and NLMR, respectively.}
\label{fig:Sim1}
\end{figure}

\begin{figure}[h!]
\centering
\subfigure[]{
   \includegraphics[scale =0.65] {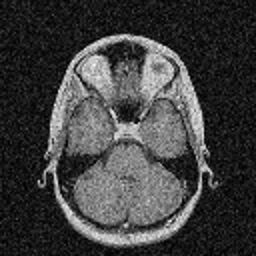}
   \label{fig:Sim2Noisy}
 }
 \subfigure[]{
   \includegraphics[scale =0.65] {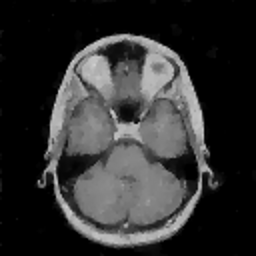}
   \label{fig:Sim2TV}
 }
\subfigure[]{
   \includegraphics[scale =0.65] {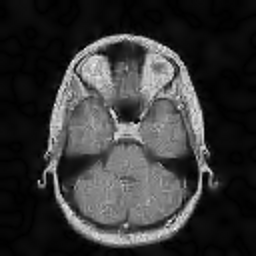}
   \label{fig:Sim2Pizurica}
 }
 \subfigure[]{
   \includegraphics[scale =0.65] {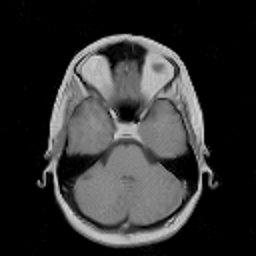}
   \label{fig:Sim2Manjon}
 }
 \subfigure[]{
   \includegraphics[scale =0.65] {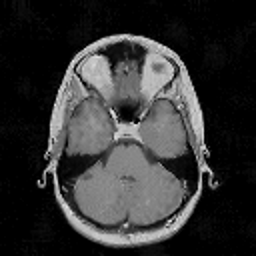}
   \label{fig:Sim2NCCS}
 }
 \subfigure[]{
   \includegraphics[scale =0.65] {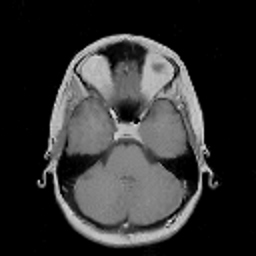}
   \label{fig:Sim2Rice}
 }
\caption{(a) Noisy version of slice \#7; (b)-(f) Reconstruction results obtained using TVDN, WDN, GNLM, NLMS and NLMR, respectively.}
\label{fig:Sim2}
\end{figure}

\begin{figure}[h!]
\centering
\subfigure[]{
   \includegraphics[scale =0.65] {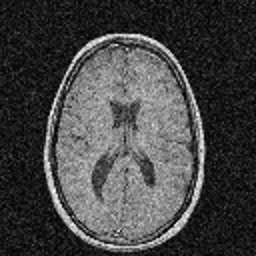}
   \label{fig:Sim2Noisy}
 }
 \subfigure[]{
   \includegraphics[scale =0.65] {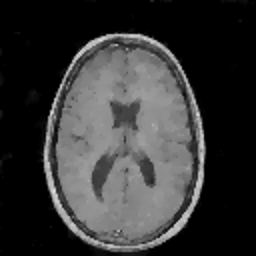}
   \label{fig:Sim2TV}
 }
\subfigure[]{
   \includegraphics[scale =0.65] {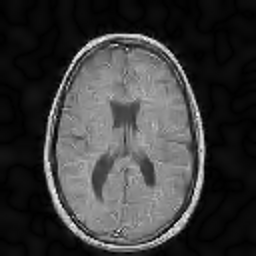}
   \label{fig:Sim2Pizurica}
 }
 \subfigure[]{
   \includegraphics[scale =0.65] {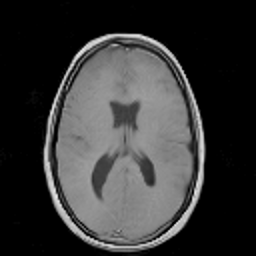}
   \label{fig:Sim2Manjon}
 }
 \subfigure[]{
   \includegraphics[scale =0.65] {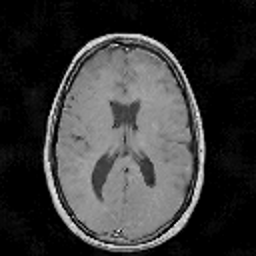}
   \label{fig:Sim2NCCS}
 }
 \subfigure[]{
   \includegraphics[scale =0.65] {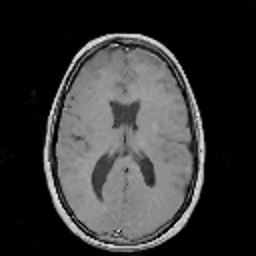}
   \label{fig:Sim4Rice}
 }
\caption{(a) Noisy version of slice \#16; (b)-(f) Reconstruction results obtained using TVDN, WDN, GNLM, NLMS and NLMR, respectively.}
\label{fig:Sim4}
\end{figure}

Table~\ref{Table:1} summarizes the values of the performance metrics of Section~\ref{PerformanceMertics} obtained for various tested images and different denoising algorithms (with the best results being accentuated with bold characters). Furthermore, for the sake of visual comparison, the image restoration results are shown in Fig.~\ref{fig:Sim1}, Fig.~\ref{fig:Sim2} and Fig.~\ref{fig:Sim4}, which have the same composition. Namely, Subplots (a) of the figures show the noisy data images, while Subplots (b)-(f) show the recovered images obtained using the TVDN, WDN, GNLM, NLMS and NLMR algorithms, respectively. 

\begin{figure}[h!]
\centering
\subfigure[]{
   \includegraphics[scale =0.65] {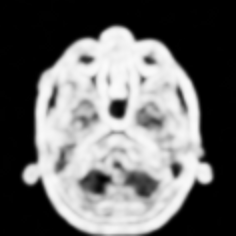}
   \label{fig:Sim1NoisyMap}
 }
 \subfigure[]{
   \includegraphics[scale =0.65] {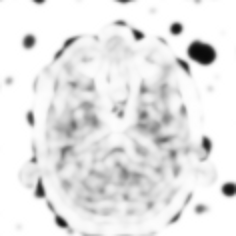}
   \label{fig:Sim1TVMap}
 }
\subfigure[]{
   \includegraphics[scale =0.65] {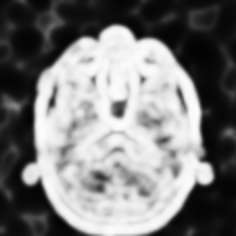}
   \label{fig:Sim1PizuricaMap}
 }
 \subfigure[]{
   \includegraphics[scale =0.65] {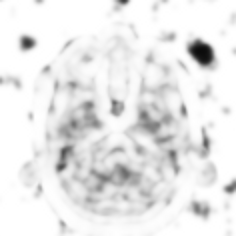}
   \label{fig:Sim1ManjonMap}
 }
 \subfigure[]{
   \includegraphics[scale =0.65] {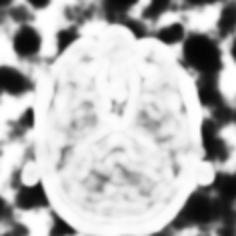}
   \label{fig:Sim1NCCSMap}
 }
 \subfigure[]{
   \includegraphics[scale =0.65] {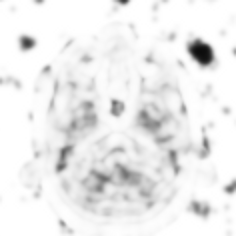}
   \label{fig:Sim1RiceMap}
 }
\caption{SSIM maps of (a) noisy version of slice \#4 and (b)-(f) reconstruction results obtained using TVDN, WDN, GNLM, NLMS and NLMR, respectively.}
\label{fig:Sim1SSIMMap}
\end{figure}

As can be seen from Table~\ref{Table:1} as well as from its related figures, the NLM algorithms produce better reconstruction results as compared to TVDN and WDN. Moreover, among the NLM algorithms, NLMR provides better performance than NLMS and GNLM in terms of all the performance measures. It is worthwhile noting that the performance of NLMS is comparable to that of NLMR and GNLM in terms of the RMSE and cRMSE measures, while the former presents significantly lower values of SSIM as compared to the other two. To further explore this phenomenon, Fig.~\ref{fig:Sim1SSIMMap} shows the SSIM maps \cite{Wang04} computed for the images of Fig.~\ref{fig:Sim1}. The SSIM maps represent the local values of SSIM (with their brighter intensities indicating stronger resemblance between the reconstructed and original MRI images), and hence they are particularly suitable for analyzing the spatial distribution of reconstruction errors. Thus, for example, Subplot (a) of Fig.~\ref{fig:Sim1SSIMMap} corresponds to the noisy data image, in which case the SSIM values at the image background are negligibly small, indicating little resemblance between the original (monotone) and measured (noisy) data. Similarly, the SSIM map corresponding to the NLMS reconstruction (as shown in Subplot (e) of Fig.~\ref{fig:Sim1}) has considerably darker background values in comparison to the GNLM and NLMR maps (shown in Subplots (d) and (f), respectively).  The NLMS reconstruction's reduced background resemblance is mainly due to the bias subtraction (thresholding) procedure discussed in Section~\ref{Sec:BiasRemoval}. Even through a visual inspection fails to find a difference between the original and the reconstructed background, the accumulation of small errors over a relatively large number of background pixels adversely affects the average value of the SSIM metric, as indicated by Table~\ref{Table:1}.

An additional important observation can be made through comparing the values of RMSE and cRMSE metrics obtained using different reconstruction algorithms. In particular, we first note that the values of RMSE and cRMSE corresponding to the noisy data are not identical -- the fact which indicates the presence of a non-zero bias in the measurement noise. At the same time, these metrics have approximately equal values for the case of TVDN, GNLM, NLMS and NLMR reconstructions, while being noticeably different in the case of WDN. This fact suggests that the latter method is inefficient in removing the constant bias in the reconstruction error.

\begin{table}
\centering
\caption{Performance metrics of the result of denoising using different algorithms}
\label{Table:1}
\begin{tabular}{|c| c | c | c  |c | c | c | c | c| c|}
\hline
Image & \multicolumn{3}{c|}{MRI Slice 4} & \multicolumn{3}{c|}{MRI Slice 7} & \multicolumn{3}{c|}{MRI Slice 16}\\
\hline\hline
 	&	RMSE	& 	cRMSE	& 	SSIM	&	RMSE	&	cRMSE	&	SSIM	&	RMSE	&	cRMSE & SSIM\\
 	\hline
Noisy 	& 	21.06 		& 	18.6175 	& 	0.4187 	& 20.8266 	& 	18.6651 	& 	0.4193 	&   20.8274		&  	18.7672 	& 	0.3732\\\hline
TVDN 	& 	13.8097 	& 	13. 8094 	& 	0.9028 &  	13.1824 	& 	13.1777		& 	0.8974 	& 	13.4104 	& 	13.4078		& 	0.8878\\	\hline
WDN 		& 	16.0125 	& 	14.9681 	& 	0.4849	& 	15.3424 	& 	14.4047 	& 	0.5158 	& 	14.9729 	& 	13.9367 	& 	0.4890\\ \hline
GNLM 	&	13.4437 	& 	13.4261 	& 	0. 9128  &	12.4910 	& 	12.4715 	& 	0.9104 &	11.3428 	& 	11.3431 	& 	0.8920\\ \hline
NLMS 	& 	13.3571 	& 	13.1606 	& 	0.7066  & 	12.5003 	& 	12.3805 	& 	0.7543 	& 	11.9151 	& 	11.6872 	& 	0.7163\\ \hline
NLMR 	& {\bf 12.6287} & {\bf 12.6115} & {\bf 0.9270}& {\bf 11.8773}	& {\bf 11.8554} &  {\bf 0.9204} & {\bf 10.6613}	& {\bf 10.6534} &  {\bf 0.9110}\\\hline
\end{tabular}
\end{table}

\subsection{Experiments with real-life data}
\begin{figure}[ht]
\centering
\subfigure[]{
   \includegraphics[scale =0.5] {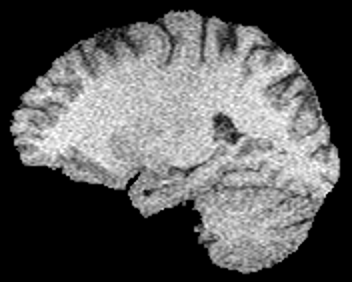}
   \label{fig:SagitalNoisy}
 }
 \subfigure[]{
   \includegraphics[scale =0.5] {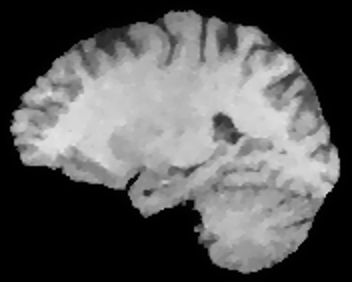}
   \label{fig:SagitalTV}
 }
\subfigure[]{
   \includegraphics[scale =0.5] {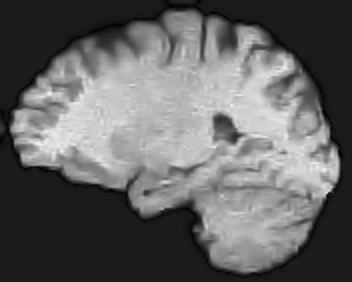}
   \label{fig:PizuricaSagital}
 }
 \subfigure[]{
   \includegraphics[scale =0.5] {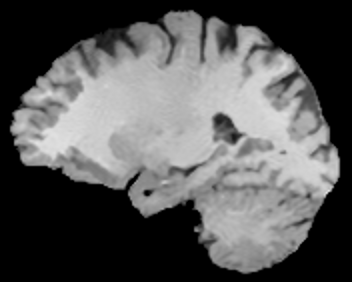}
   \label{fig:ManjonSagital}
 }
 \subfigure[]{
   \includegraphics[scale =0.5] {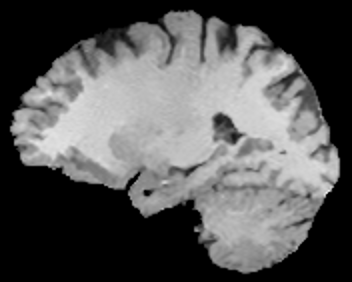}
   \label{fig:ProposedRicianSagital}
 }
 \subfigure[]{
   \includegraphics[scale =0.5] {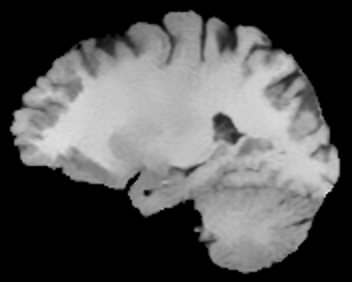}
   \label{fig:ProposedRicianSagital}
 }
\caption{(a) Sagittal MRI scan; (b)-(f) Reconstruction results obtained using TVDN, WDN, GNLM, NLMS and NLMR, respectively.}
\label{fig:Sagital}
\end{figure}

\begin{figure}[ht]
\centering
\subfigure[]{
   \includegraphics[scale =0.5] {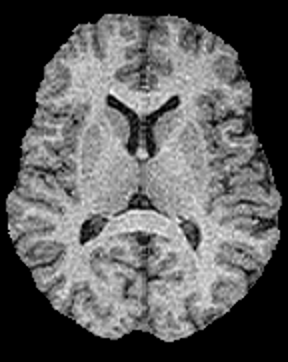}
   \label{fig:AxialNoisy}
 }
 \subfigure[]{
   \includegraphics[scale =0.5] {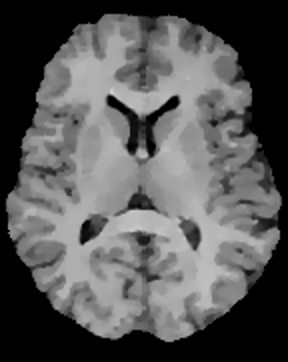}
   \label{fig:AxialTV}
 }
\subfigure[]{
   \includegraphics[scale =0.5] {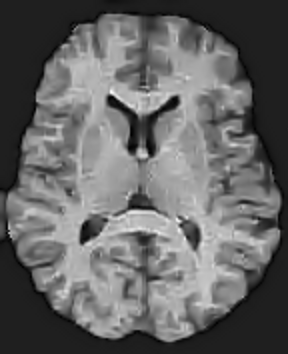}
   \label{fig:AxialPizurica}
 }
 \subfigure[]{
   \includegraphics[scale =0.5] {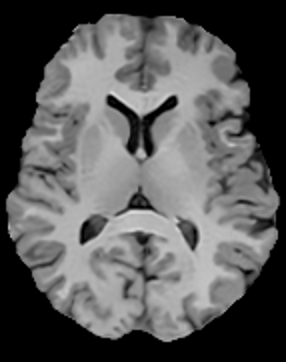}
   \label{fig:AxialManjon}
 }
 \subfigure[]{
   \includegraphics[scale =0.5] {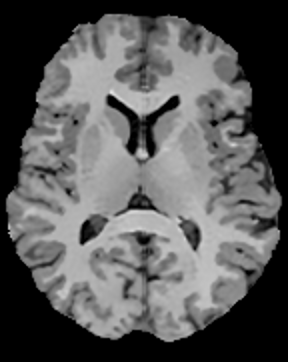}
   \label{fig:AxialRice}
 }
 \subfigure[]{
   \includegraphics[scale =0.5] {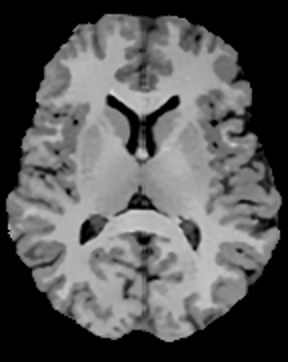}
   \label{fig:AxialRice}
 }
\caption{(a) Axial MRI scan; (b)-(f) Reconstruction results obtained using TVDN, WDN, GNLM, NLMS and NLMR, respectively.}
\label{fig:Axial}
\end{figure}

Reconstruction of real-life MRI images has been the next step in our comparative study. To this end, the data set of \cite{Pizurica03} have been used herein. The data were obtained at the University Hospital of Ghent and it is publicly available at {\tt http://telin.ugent.be/\~{}sanja/Sanja\_files/Software/MRIprogram.zip}. The data contains a sagittal and an axial scan of a human brain, which are shown in Fig.~\ref{fig:SagitalNoisy} and Fig.~\ref{fig:AxialNoisy}, respectively. 

The reconstruction results obtained for each of the tested images using the proposed and reference methods are shown in Subplots (b)-(f) of Fig.~\ref{fig:Sagital} and Fig.~\ref{fig:Axial}, respectively. From these figures, it can be seen that the proposed algorithms result in higher-contrast reconstructions of better visual clarity as compared to the reference approaches. The difference is particularly evident for the case of Fig.~\ref{fig:Axial}, where the proposed algorithms result in less noisy images, while exhibiting higher effective resolution and contrast.

\section{Conclusion}\label{Sec:Discussion}
The present paper has proposed two novel NLM-based methods for enhancement of MR images. More specifically, the paper introduced a new definition of the NLM weights, which takes into consideration the true Rician statistics of measurement noises. The proposed definition has been shown to overcome some unfavourable characteristics of the weights previously proposed in the MRI denoising literature. Subsequently, the current paper provided closed form expressions for the weights corresponding to both subtractive and rational similarity measures. The utility of the proposed algorithm has been demonstrated through a series of computer simulations and real-life experiments. Based on the obtained results, one can find that the proposed algorithm has been able to provide better reconstruction results as compared to a number of established reference approaches.

\begin{appendices}
\section{Subtractive SM for non-central chi square statistics}\label{App:NCCS}
We are interested to evaluate ${\rm SSM}=p_{F_{s-k}-F_{t-k}\mid G_{s-k},G_{t-k}}(0\mid g_{s-k},g_{t-k})$, which is defined as \cite{Teuber11}
\begin{equation}
{\rm SSM}=\int_{0}^\infty p_{F_{s-k}\mid G_{s-k},G_{t-k}}(f\mid g_{s-k},g_{t-k}) \,  p_{F_{t-k}\mid G_{s-k},G_{t-k}}(f\mid g_{s-k},g_{t-k})\,df,
\end{equation}
where it has been assumed that $F_{s-k}$ and $F_{t-k}$ are conditionally independent, given the noisy data $G_{s-k}$ and $G_{t-k}$. Moreover, assuming
\begin{equation}
p_{F_{z_1-k} \mid G_{z_1-k}, G_{z_2-k}}(f|g_{z_1-k},g_{z_2-k}) =  p_{F_{z_1-k} \mid G_{z_1-k}}(f|g_{z_1-k}),~~~z_1,z_2\in \{s,t\}
\end{equation}
and using Bayes theorem, one obtains 
\begin{equation}
{\rm SSM}=\frac{\int_{0}^\infty p_{F_{s-k}}(f) \, p_{F_{t-k}}(f) \, p_{G_{s-k}\mid F_{s-k}}(g_{s-k}\mid f) \,p_{G_{t-k}\mid F_{t-k}}(g_{t-k}\mid f)\,df}{p_{G_{s-k}}(g_{s-k}) \, p_{G_{t-k}}(g_{t-k})}.
\end{equation}
If no prior information about the original intensities is known, one can assume that $F_{s-k}$ and $F_{t-k}$ follow a uniform distribution \cite{Deledalle09,Teuber11}. Also, for the case of an NCCS distribution, $p_{G_{z-k}}(g_{z-k})$, where $z\in \{s,t\}$, can be written as
\begin{align}\label{eq:pG}
p_{G_{z-k}}(g_{z-k})&=\int_{0}^\infty p_{G_{z-k}\mid F_{z-k}}(g_{z-k}\mid f_{z-k}) \, p_{F_{z-k}}(f)\,df = \notag\\
&=p_{F_{z-k}}(f)\int_{0}^\infty \frac{1}{2} e^{-(g_{z-k}+f)/2}I_0(\sqrt{fg_{z-k}})\,df = \\
&=p_{F_{z-k}}(f), \notag
\end{align}
where the second line in \eqref{eq:pG} follows from the uniform density assumption of $F_{z-k}$ and the integration result is obtained by noting that $p_{G_{z-k}\mid F_{z-k}}$ can also be looked upon as a probability density function of $F_{z-k}$ with $g_{z-k}$ as its parameter. Hence, in the case of NCCS noise distribution, the expression for ${\rm SSM}$ is given by 
\begin{align}
{\rm SSM}=\MS&=\int_{0}^\infty p_{G_{s-k}\mid F_{s-k}}(g_{s-k}\mid f) \, p_{G_{t-k}\mid F_{t-k}}(g_{t-k}\mid f)\,df\notag\\
&=\frac{1}{4}e^{-(g_{s-k}+g_{t-k})/2}\int_{0}^\infty e^{-f} I_0(\sqrt{g_{s-k}f}) \, I_0(\sqrt{g_{t-k}f})\,df.
\end{align}
Substituting $y=\sqrt{f}$ results in
\begin{align}
\MS
&=\frac{1}{2}e^{-(g_{s-k}+g_{t-k})/2}\int_0^\infty y\, e^{-y^2}I_0(\sqrt{g_{s-k}}y) \, I_0(\sqrt{g_{t-k}}y)\,dy = \notag\\
&=\frac{1}{2}e^{-(g_{s-k}+g_{t-k})/2}\int_0^\infty y \, e^{-y^2} \, J_0(\imath \sqrt{g_{s-k}}y) \, J_0(\imath \sqrt{g_{t-k}}y)\,dy,
\end{align}
where $J_0$ is the zero-order Bessel function of the first kind, and $\imath = \sqrt{-1}$. Subsequently, using equation (2.32) of \cite{Lawrence94}, we have
\begin{align}\label{eq:NCCSmeasure}
\MS
&=\frac{1}{4}e^{-(g_{s-k}+g_{t-k})/4}I_0\left(-\frac{\sqrt{g_{s-k}g_{t-k}}}{2}\right) = \notag\\
&=\frac{1}{4}e^{-(g_{s-k}+g_{t-k})/4}I_0\left(\frac{\sqrt{g_{s-k}g_{t-k}}}{2}\right),
\end{align}
where the last line stems from the fact that $I_0(-x)=I_0(x)$.

\section{Rational SM for Rician statistics}\label{App:Rician}
In the case of Rician statistics, the rational similarity measure is given by
\begin{align}
\MR &=p_{A_{s-k}/A_{t-k}\mid M_{s-k},M_{t-k}}(1\mid m_{s-k},m_{t-k}) = \notag\\
&=\int_{0}^\infty a \, p_{A_{s-k}\mid M_{s-k}}(a\mid m_{s-k}) \, p_{A_{t-k}\mid M_{t-k}}(a\mid m_{t-k})\,da.
\end{align}
Proceeding with a similar derivation as in Appendix \ref{App:NCCS}, we obtain
\begin{align}\label{eq:AppSNL2}
\MR&= \int_{0}^\infty a \, p_{M_{s-k}\mid A_{s-k}}(m_{s-k}\mid a) \, p_{M_{t-k}\mid A_{t-k}}(m_{t-k}\mid a)\,da = \notag\\
&=\frac{m_{s-k}m_{t-k}}{\sigma^4}e^{-(m_{s-k}^2+m_{t-k}^2)/2\sigma^2}\int_0^\infty a \, e^{-{a^2}/{\sigma^2}} I_0\left(\frac{m_{s-k}a}{\sigma^2}\right) \, I_0\left(\frac{m_{t-k}a}{\sigma^2}\right)\,da.
\end{align}
Substituting $y=a/\sigma$ in \eqref{eq:AppSNL2} results in
\begin{align}
\MR &=\frac{m_{s-k}m_{t-k}}{\sigma^2}e^{-(m_{s-k}^2+m_{t-k}^2)/2\sigma^2}\int_{0}^\infty y \, e^{-y^2} I_0\left(\frac{m_{s-k}}{\sigma}y\right) I_0\left(\frac{m_{t-k}}{\sigma}y\right)\,dy.
\end{align}
Finally, using formula (2.32) of \cite{Lawrence94}, we have
\begin{equation}
\MR=
\frac{m_{s-k}m_{t-k}}{2\sigma^2}e^{-(m_{s-k}^2+m_{t-k}^2)/4\sigma^2} I_0\left(\frac{m_{s-k}m_{t-k}}{2\sigma^2}\right).
\end{equation}
\end{appendices}

\bibliographystyle{IEEEtran}
\bibliography{IEEEabrv,refers}

\end{document}